\documentclass[10pt,twocolumn,letterpaper]{article}

\usepackage[pagenumbers]{iccv} 

\definecolor{iccvblue}{rgb}{0.21,0.49,0.74}
\usepackage[pagebackref,breaklinks,colorlinks,allcolors=iccvblue]{hyperref}
\usepackage{multirow}
\usepackage{multicol}
\usepackage{hyperref}
\usepackage{xcolor}
\usepackage{multirow}
\usepackage{multicol}
\usepackage{colortbl}
\usepackage{verbatim}
\usepackage{graphicx}
\usepackage{booktabs}
\usepackage{multirow}
\usepackage{listings}
\usepackage{xcolor}
\usepackage[T1]{fontenc}
\def\paperID{3375} 
\def\confName{ICCV}
\def\confYear{2025}

\title{Referring to Any Person}

\author{Qing Jiang$^{1,2}$ , Lin Wu$^{1,2}$ , Zhaoyang Zeng$^{1}$ , Tianhe Ren$^{1}$ , Yuda Xiong$^{1}$ \\ Yihao Chen$^{1}$ , Liu Qin$^{1}$ , Lei Zhang$^{1,2\dagger}$ \\
$^1$International Digital Economy Academy (IDEA) \\
$^2$South China University of Technology \\ 
{\tt\small mountchicken@outlook.com , leizhang@idea.edu.cn} \\
\url{https://deepdataspace.com/blog/dino-xseek}
}

\begin{document}
\twocolumn[{
\maketitle\centering
\captionsetup{type=figure}
\includegraphics[width=0.98\textwidth]{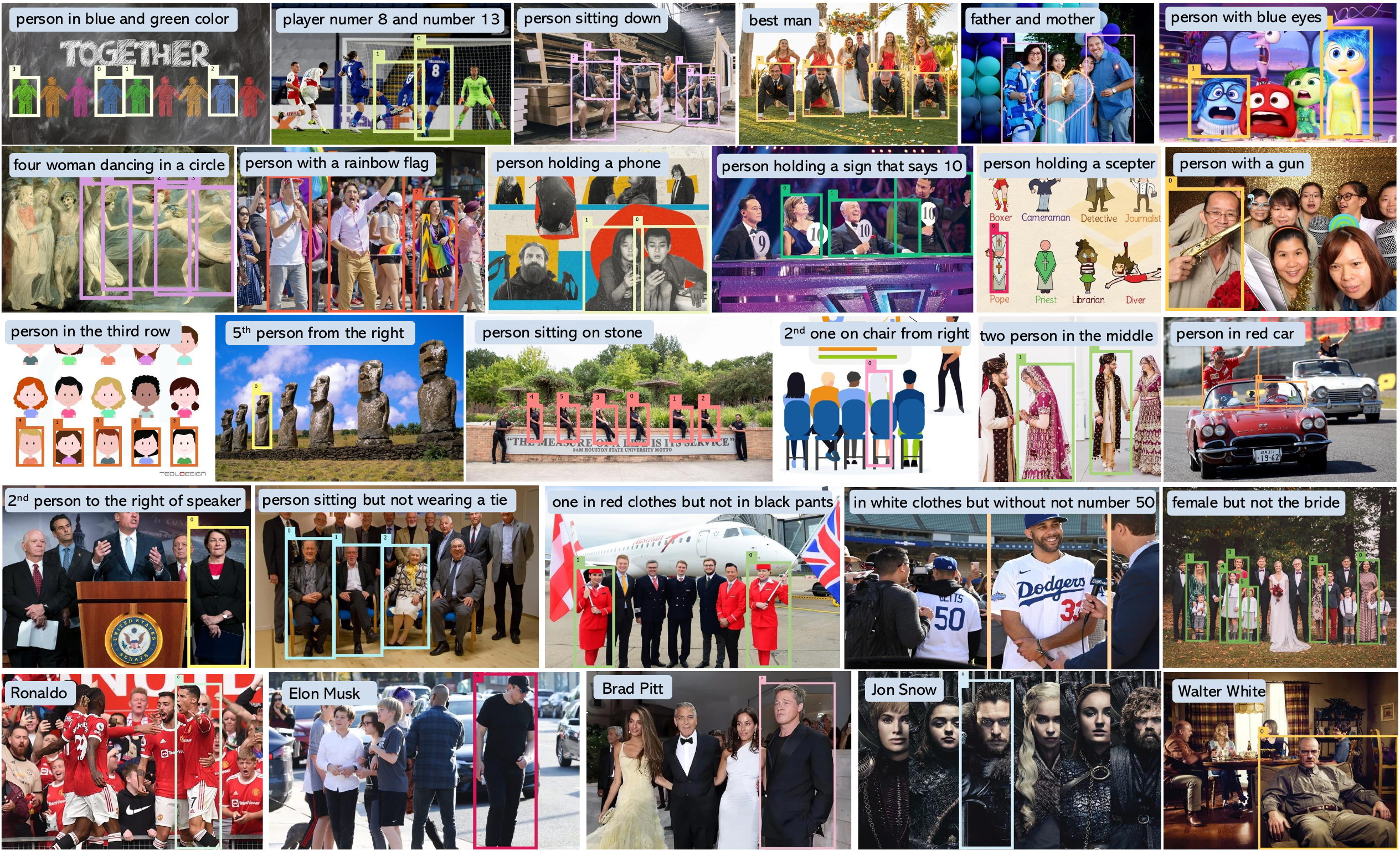}
\captionof{figure}{We introduce referring to any person, a task that requires detecting all individuals in an image which match a given natural language description, and a new model RexSeek designed for this task with strong perception and understanding capabilities that effectively captures attributes, spatial relations, interactions, reasoning, celebrity recognition, etc.}
\label{fig:teaser}
}]
\makeatletter
\def\@makefnmark{} 
\makeatother
\footnotetext{This work was done when Qing Jiang and Lin Wu were interns at IDEA.}

\footnotemark
\addtocounter{footnote}{-1}
\footnotetext{\emph{$\dagger$ Corresponding author.}}

\maketitle
\begin{abstract}
Humans are undoubtedly the most important participants in computer vision, and the ability to detect any individual given a natural language description, a task we define as referring to any person, holds substantial practical value. However, we find that existing models generally fail to achieve real-world usability, and current benchmarks are limited by their focus on one-to-one referring, that hinder progress in this area. In this work, we revisit this task from three critical perspectives: task definition, dataset design, and model architecture. We first identify five aspects of referable entities and three distinctive characteristics of this task. Next, we introduce HumanRef, a novel dataset designed to tackle these challenges and better reflect real-world applications. From a model design perspective, we integrate a multimodal large language model with an object detection framework, constructing a robust referring model named RexSeek. Experimental results reveal that state-of-the-art models, which perform well on commonly used benchmarks like RefCOCO/+/g, struggle with HumanRef due to their inability to detect multiple individuals. In contrast, RexSeek not only excels in human referring but also generalizes effectively to common object referring, making it broadly applicable across various perception tasks. Code is available at \url{https://github.com/IDEA-Research/RexSeek}

\end{abstract}    
\section{Introduction}
\label{seuc:intro}

Humans are central to computer vision~\cite{khirodkar2024sapiens,tang2023humanbench,chen2023beyond,ci2023unihcp,ju2023humansd,ju2023human,chen2023humanmac,lu2023humantomato,chen2017beyond,toshev2014deeppose,loper2023smpl,bogo2016keep,guler2019holopose,zheng2019deephuman, zheng2017person}, and the ability to identify and detect specific individuals based on natural language descriptions, a task we define as referring to any person, is crucial for numerous applications, including human-robot interaction, industrial automation, healthcare, etc.

\begin{figure}[t]
\centering
\includegraphics[width=1.0\linewidth]{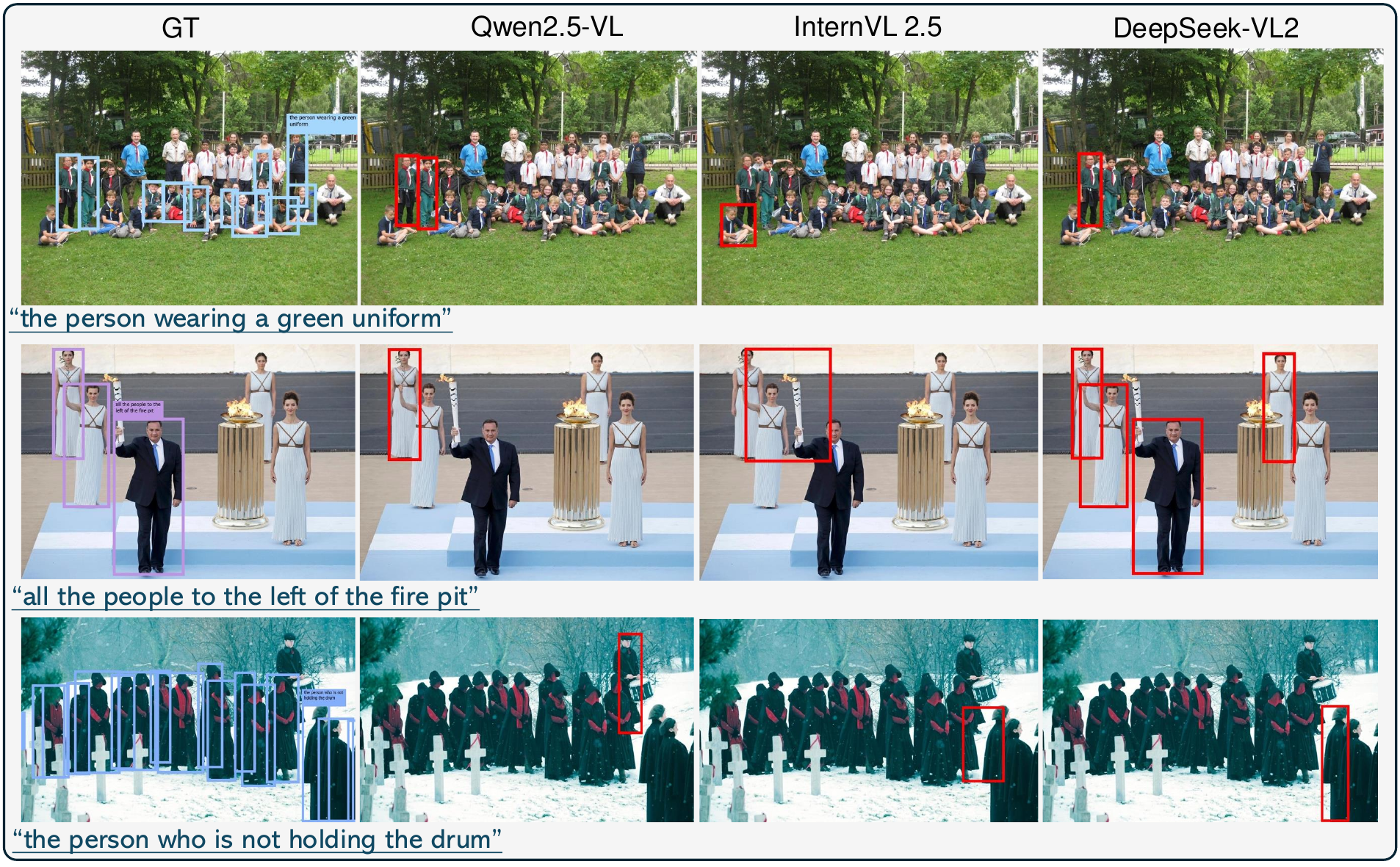}\vspace{-1mm}
\caption{Visualization results of Qwen2.5-VL~\cite{bai2025qwen2}, InternVL-2.5~\cite{chen2024expanding}, and DeepSeek-VL2~\cite{wu2024deepseek} on the human referring task. Despite achieving strong results on referring benchmarks RefCOCO/+/g~\cite{Datasets:REFCOCOG, Datasets:REFCOCO}, state-of-the-art models struggle when tasked with identifying multiple individuals as they output an insufficient number of bounding boxes.}
\label{fig:intro1} 
\vspace{-4mm}
\end{figure}

However, we argue that progress in this area has been hindered by unclear task definitions and a lack of high-quality data. Our findings show that despite achieving state-of-the-art performance on referring benchmarks RefCOCO/+/g~\cite{Datasets:REFCOCOG, Datasets:REFCOCO}, most models remain impractical for real-world applications, as illustrated in Figure \ref{fig:intro1}. To address this challenge, we revisit this task from three perspectives: task definition, dataset construction, and model design.

We begin by formally defining the task of referring to any person: \textit{given a natural language description and an input image, the model needs to detect all individuals in the image who match the description}. To comprehensively capture the scope of this task, we identify five key aspects that define how humans can be referred to:
\textbf{i) Attributes: }Encompassing intrinsic characteristics such as gender, age, action, clothing, etc.
\textbf{ii) Position: }Describing spatial relationships both among individuals and between individuals and their surroundings.
\textbf{iii) Interaction: }Accounting for human-to-human, human-to-object, and human-to-environment interactions.
\textbf{iv) Reasoning: }Involving multi-step inference that considers multiple objects to resolve complex referrings.
\textbf{v) Celebrity Recognition: }Identifying specific individuals, whether by their real names or characters names.

Next, we identify three crucial characteristics that define this task:
\textbf{i) Multi-Instance Referring}: A referring expression can correspond to multiple individuals. While mainstream referring datasets RefCOCO/+/g~\cite{Datasets:REFCOCOG, Datasets:REFCOCO} typically assume that each expression refers to a single object, this does not align with real-world scenarios. We find through experiments that most models experience significant performance degradation when tasked with identifying more than one individual. \textbf{ii) Multi-Instance Discrimination}: The image should contain multiple individuals in addition to the target person. This setting ensures that the model fully comprehends the referring expression to identify the correct individual rather than simply detecting all people in the image.
\textbf{iii) Rejection of Non-existence}: If the referred person is not present in the image, the model should refuse to generate a result rather than produce a hallucinated output.

Based on the task definition, we manually constructed a novel dataset for human referring, named HumanRef. Unlike the traditional ReferItGame~\cite{kazemzadeh2014referitgame} annotation approach, where one annotator describes an object and another finds it based on the description, we adopt a different annotation methodology. Our process begins with annotators listing the key properties of individuals in an image according to the predefined referable entities. Next, for each person, they determine whether these properties apply and result in a property dictionary. Finally, a large language model~\cite{yang2024qwen2} composes these properties into referring expressions. HumanRef comprises 103,028 referring statements, with each expression referring to an average of 2.2 instances.  We also split a benchmark from HumanRef with 6,000 referring expressions spanning six subsets, ensuring comprehensive coverage across all referable properties.

From the model design perspective, we argue that a robust referring model should possess two key characteristics: 
\textbf{i) Robust Perception Ability}: The model should be capable of detecting all individuals in an image. \textbf{ii) Strong Language Comprehension}: The model should effectively interpret complex language descriptions of people. To address these requirements, we introduce RexSeek, a detection-oriented multimodal large language model specifically designed for this task. Inspired by ChatRex~\cite{jiang2024chatrex}, we formulate referring as a retrieval-based task. RexSeek lintegrates a person detector~\cite{ren2024dino} as its box input, ensuring strong perception capabilities while incorporating Qwen2.5~\cite{yang2024qwen2} as the LLM to enhance language comprehension. We adopt a multi-stage training approach that progressively refines both detection and comprehension skills, equipping RexSeek with strong referring capabilities.

Experimental results indicate that most state-of-the-art models~\cite{you2023ferret, ma2024groma, chen2023shikra, bai2025qwen2, wu2024deepseek, chen2024expanding, ren2024dino, jiang2024chatrex} exhibit performance degradation on the HumanRef benchmark, despite achieving strong results on RefCOCO/+/g. The primary limitation is that these models typically detect only a single instance, as they are trained on datasets that assume one-to-one referring. In contrast, RexSeek, trained on HumanRef, exhibits strong referring capabilities. Additionally, benefiting from the multi-stage training approach, RexSeek also emerges with the ability to refer to generalized objects, extending its applicability beyond human-centric tasks. To summarize, our contributions are threefold:

\begin{itemize}
    \item We introduce referring to any person with a clear definition by identifying five aspects of referable entities and three key characteristics that distinguish this task.
    \item We introduce HumanRef, a novel referring dataset, and establish a challenging benchmark to drive progress in human-centric referring expression research.
    \item We propose RexSeek, a detection-oriented multimodal large language model trained through a multi-stage process, demonstrating strong referring capabilities for both humans and general objects.
\end{itemize}

\section{Related Work}
\label{seuc:related_work}
\textbf{Referring Expression Comprehension Task.}
Referring Expression Comprehension (REC)~\cite{qiao2020referring,Datasets:REFCOCO,Datasets:REFCOCOG,kazemzadeh2014referitgame,zhang2019referring,luo2020multi,yu2018mattnet,yu2018mattnet,yang2019dynamic,liao2020real} involves interpreting a natural language expression to localize specific objects within an image. Unlike open-vocabulary object detection~\cite{jiang2025t, ren2024dino, ren2024grounding, liu2024grounding, yao2022detclip, li2022grounded,cheng2024yolo,minderer2022simple,zareian2021open,wu2023aligning} or phrase grounding~\cite{plummer2015flickr30k, wu2020phrasecut, krishna2017vg,gupta2020contrastive,datta2019align2ground}, which identify objects based on brief category names or short phrases, REC requires understanding complex, free-form descriptions. This task necessitates not only recognizing object attributes and relationships but also comprehending spatial configurations and interactions, making it inherently more challenging. In this work, we systematically analyze the referable entities and the critical characteristics that define this task.

\textbf{REC Datasets and Benchmarks.}
The first large-scale Referring Expression Comprehension (REC) dataset, ReferItGame~\cite{kazemzadeh2014referitgame}, was created through a two-player game in which one annotator describes an object, and another selects it. This was later followed by more sophisticated datasets~\cite{plummer2015flickr30k,de2017guesswhat,chen2020cops,chen2019touchdown,cirik2020refer360,qi2020reverie}, such as RefCOCO~\cite{Datasets:REFCOCO}, RefCOCO+~\cite{Datasets:REFCOCO}, and RefCOCOg~\cite{Datasets:REFCOCOG}, which leverage MSCOCO~\cite{Datasets:MSCOCO} images to provide more complex referring expressions. Beyond these general datasets, others address specific challenges. CLEVR-Ref+~\cite{liu2019clevr} focuses on geometric object referring. RefCrowd~\cite{qiu2022refcrowd} targets person detection in crowded scenes. Ref-L4~\cite{chen2024revisiting} handles longer and more detailed descriptions. GRES~\cite{yu2024revisiting} introduces multi-target referring expression segmentation. However, existing datasets typically assume a one-to-one correspondence between a referring expression and a single instance, which fails to reflect real-world scenarios. To address this gap, we refine the referring task and introduce HumanRef, a dataset specifically designed to support multi-instance referring and advance research in this domain.

\textbf{MLLM-based REC Methods}
Multimodal Large Language Models (MLLMs)~\cite{gpt4v, bai2025qwen2, wu2024deepseek, chen2024expanding, alayrac2022flamingo, agrawal2024pixtral, li2024aria, deitke2024molmo, wang2024qwen2,lu2024ovis,zhang2024mm1,tong2024cambrian,li2024llava,li2025eagle,liu2024nvila} have demonstrated strong capabilities in both text and image comprehension, motivating efforts to integrate referring expression understanding into these models. A common approach involves outputting bounding box coordinates as tokens~\cite{chen2023shikra,you2023ferret,zhang2024ferret,wang2023cogvlm,zhan2025griffon,zhan2024griffon,bai2025qwen2, wu2024deepseek, chen2024expanding, VLM:MM1}. Alternatively, methods like Groma~\cite{ma2024groma} and ChatRex~\cite{jiang2024chatrex} frame detection as a retrieval task, where a proposal model generates bounding boxes, and the LLM selects the index of the relevant box based on the referring expression. While these MLLM-based methods achieve high performance on RefCOCO/+/g, our experiments reveal that they remain inadequate for practical applications due to low recall rate on multi-instance referrings.
\section{HumanRef Dataset}
\label{seuc:dataset}

\begin{figure*}[h]\centering
\includegraphics[width=1\linewidth]{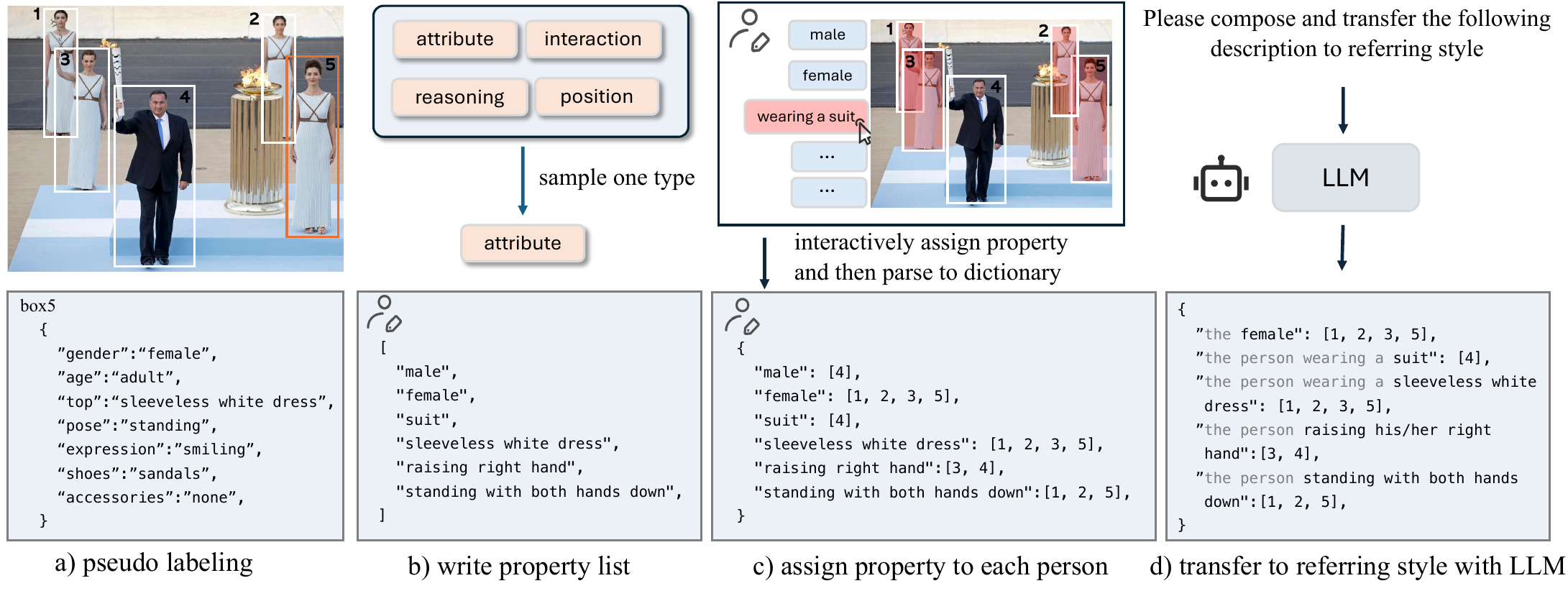}\vspace{-1mm}
\caption{Overview of the mannual annotation pipeline of the HumanRef dataset. }
\label{fig:HumanRef}
\vspace{-1mm}
\end{figure*}

In this section, we present the design philosophy, data acquisition process, annotation pipeline, and dataset statistics of the proposed HumanRef dataset.
\begin{table}[]
  \centering
  \resizebox{1\columnwidth}{!}{%
   \begin{tabular}{cll}
\hline
domain                                                          & \multicolumn{1}{c}{sub-domains}                                                                                                     & \multicolumn{1}{c}{examples}                                                                                                                                                           \\ \hline
attribute                                                        & \begin{tabular}[c]{@{}l@{}}gender, age,  race,  profession,  posture, \\ appearance,  clothing and accessories, action\end{tabular} & \textit{\begin{tabular}[c]{@{}l@{}}male, female, white man, the police officer,\\ person with a shocked expression, person wearing\\ a mask, person standing\end{tabular}}             \\ \hline
position                                                        & \begin{tabular}[c]{@{}l@{}}inner position (human to human),\\ outer position (human to environment)\end{tabular}                    & \textit{\begin{tabular}[c]{@{}l@{}}the second person from left to right, person at\\ the right, person closest to the microphone, \\ person sitting in the chair\end{tabular}}         \\ \hline
interaction                                                     & \begin{tabular}[c]{@{}l@{}}inner interaction (human with human),\\ outer interaction (human with environment)\end{tabular}          & \textit{\begin{tabular}[c]{@{}l@{}}two people holding hands,  people locked in\\ each other's gaze,  the person holding a gun, \\ person holding the certificate in hand\end{tabular}} \\ \hline
reasoning                                                       & \begin{tabular}[c]{@{}l@{}}inner position reasoning,\\ outer position reasoning,\\ attribute reasoning\end{tabular}                                                    & \textit{\begin{tabular}[c]{@{}l@{}}all the people to the right of the person closest\\ to the glass, person wearing a lab coat but not putting\\ their hand on the board\end{tabular}} \\ \hline
\begin{tabular}[c]{@{}c@{}}celebrity\\ recognition\end{tabular} & \begin{tabular}[c]{@{}l@{}}actor, character, athlete, entrepreneur,\\ scientist, politician, singer\end{tabular}                    & \textit{\begin{tabular}[c]{@{}l@{}}Brad Pitt,  Bruce Wayne,  Cristiano Ronaldo, \\ Rihanna, Elon Musk,  Albert Einstein, Donald Trump\end{tabular}}                                    \\ \hline
rejection                                                       & \multicolumn{1}{l}{attribute, position, interaction, reasoning}                                                                      & \textit{a man in red hat, three women in a circle}                                                                                                                                     \\ \hline
\end{tabular}}
\caption{
The primary annotation domains and their corresponding sub-domains within HumanRef.}
\label{tab:philosphy}
\vspace{-1.5em}
\end{table}

\subsection{Data Design Philosophy}
We define five key aspects that determine how humans can be referred to using natural language, including attribute, position, interaction, reasoning, and celebrity recognition. These categories are further elaborated with definitions and examples in Table \ref{tab:philosphy}. A key distinction between HumanRef and existing referring datasets is its focus on multi-instance referring rather than one-to-one object referring. Our dataset ensures that a single referring expression can correspond to multiple individuals, providing a more realistic and practical reflection of real-world scenarios.

\subsection{Data Acquisition}
The HumanRef dataset is designed to capture human presence across diverse contexts, including natural environments, industrial settings, healthcare, sports, films, animations, etc. To ensure dataset diversity, we sourced images containing humans from web image dataset~\cite{kakaobrain2022coyo-700m}. To filter candidate images, we first retained those with a resolution larger than 1000 × 1000 pixels to ensure high-quality content. Next, we use an open-set object detector DINO-X~\cite{ren2024dino} to detect human instances. To align with the multi-instance discrimination requirement, we retain only images containing at least four individuals.

To assist the annotator in writing properties, we prompt the QwenVL-2.5~\cite{bai2025qwen2} model to create a structured property dictionary for each person in the image, capturing details such as gender, clothing, actions, etc. Ultimately, this phase produced image, person box, and person description triples used for further annotation.

\subsection{Manual Annotation}
For attribute, position, interaction, and reasoning subsets, we adopt manual annotation. This annotation process consists of three main steps, including property listing, property assignment, and referring style rewriting. Given an image, along with the corresponding person boxes and pre-labeled property dictionary, the annotation system will randomly select one annotation type from attribute, location, interaction, and reasoning to assign to the annotator. The following annotation process is then carried out:

\textbf{Property Listing:} 
The annotator examines all individuals in the image, considering both their visual appearance, action, position, interaction, and the pre-labeled property dictionary. Based on these observations, the annotator compiles a list of properties. To enhance dataset richness, annotators are encouraged to label attributes shared by multiple individuals while avoiding those common to all. Additionally, we monitor the word frequency of labeled referring expressions and restrict the use of high-frequency words to improve data diversity.
 
\textbf{Property Assignment:} 
Once the properties are listed, annotators systematically assign them to the corresponding individuals. This interactive process involves selecting a property value and clicking on the associated bounding boxes to link it to correct persons. The final output is a structured dictionary, where keys represent property names and values contain lists of bounding box indices corresponding to the individuals possessing each property.

\textbf{Referring Style Rewriting: }In the final step, we prompt Qwen2.5~\cite{yang2024qwen2} to reformulate the structured attribute dictionary into short, natural language referring expressions. The final annotated data also undergoes a thorough review process to ensure its quality. 

\subsection{Automatic Annotation}
For celebrity recognition and rejection referring, we employ two efficient and effective automatic annotation pipelines.

\begin{figure*}[h]\centering
\includegraphics[width=1\linewidth]{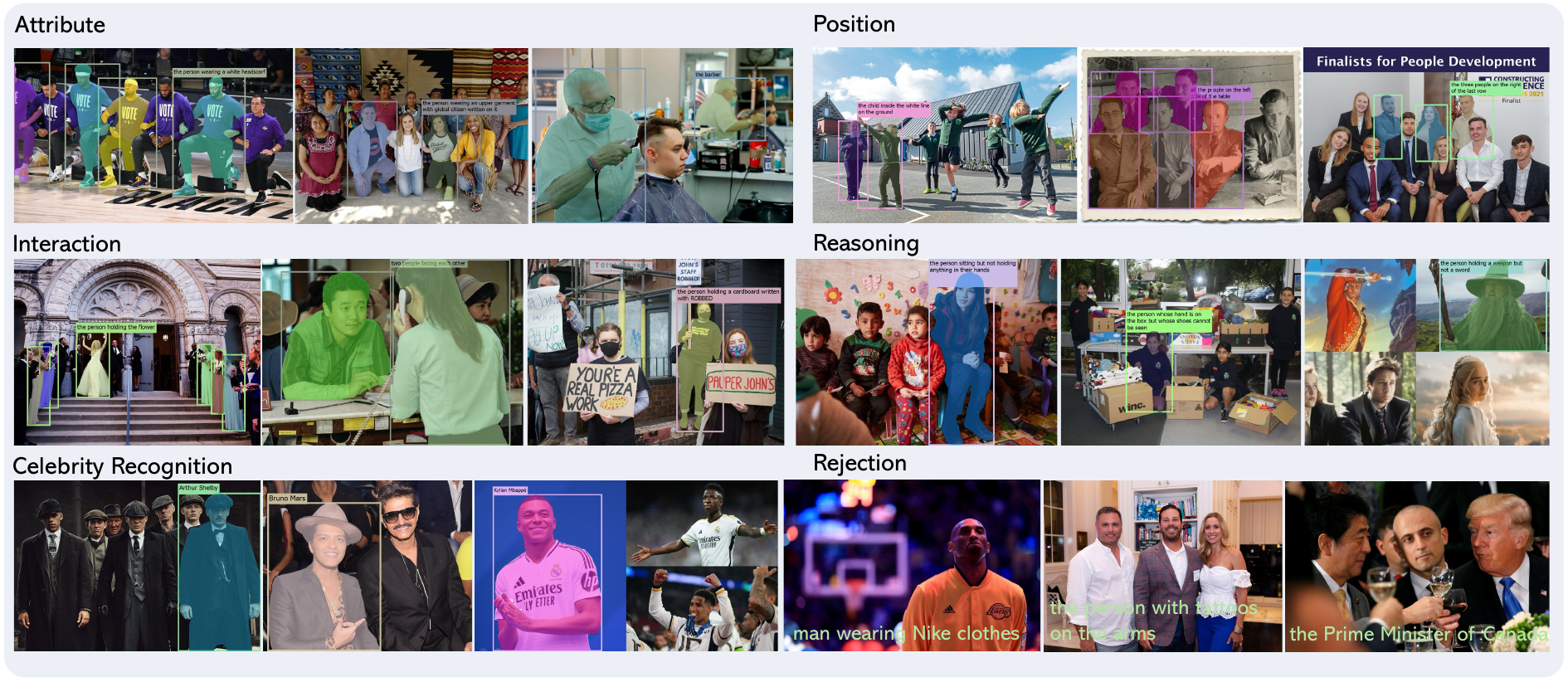}\vspace{-1mm}
\caption{Visualization of the six subsets in the HumanRef Benchmark.}
\label{fig:HumanRef_gallery}
\vspace{-2mm}
\end{figure*}

\textbf{Celebrity Recognition:} We first categorize celebrities into seven distinct fields: actors, film characters, athletes, singers, entrepreneurs, scientists, and politicians. For each field, we identify the most well-known individuals, compiling a final list of 636 names, which we then used as prompts to retrieve images via the Bing Search API\footnote{\href{https://www.microsoft.com/en-us/bing/apis}{https://www.microsoft.com/en-us/bing/apis}}. The collected images include both individual and group photos, necessitating a method to accurately associate each celebrity name with the correct person in the image. To achieve this, we first use the DINO-X~\cite{ren2024dino} model to detect all human faces and persons, linking each detected face to its corresponding person box based on overlap measurements. If an image contains only one person, we assume this individual is the target celebrity. For images featuring multiple individuals, we use a Python face recognition library\footnote{\href{https://github.com/ageitgey/face_recognition}{https://github.com/ageitgey/facerecognition}}, leveraging a single-person image as a recognition template to match and identify the same person in such images. 

\textbf{Rejection Referring:} The objective of this sub-dataset is to ensure that when a referring description targets a person who does not exist in the input image, the model rejects the referring request instead of hallucinating and outputting an incorrect bounding box.  To construct this dataset, we first extract referring expressions from the attribute, position, interaction, and reasoning subsets. We then prompt Qwen2.5~\cite{yang2024qwen2} to modify these descriptions, transforming them into similar but semantically altered versions. For instance, a description such as \textit{``the person wearing a blue hat''} may be changed to \textit{``the person wearing a red hat''}. To validate the generated descriptions, we prompt Molmo~\cite{deitke2024molmo} to detect the modified referring expression. If no matching object is found in the output, the data is retained.

\subsection{HumanRef Benchmark}
To construct the HumanRef Benchmark, we sample 1,000 referring expressions from each of the four manually annotated subsets. Additionally, for the celebrity and rejection subsets, we conduct a separate manual annotation process to create 1,000 new referring expressions for each category, ensuring high-quality and challenging evaluation data. To further support advancements in referring expression segmentation, we utilize SAM2~\cite{ravi2024sam} to generate masks for each ground truth bounding box. Figure \ref{fig:HumanRef_gallery} presents example cases from the HumanRef Benchmark, illustrating the diversity and complexity of the dataset.

\begin{table}[]
  \centering
  \resizebox{1\columnwidth}{!}{%
   \begin{tabular}{lccccccc}
\hline
\multicolumn{8}{c}{HumanRef Train}                                                                                                      \\ \hline
\multicolumn{1}{l|}{type}           & attribute & position & interaction & reasoning & celebrity & \multicolumn{1}{c|}{rejection} & total \\ \hline
\multicolumn{1}{l|}{images}         & 8,614    & 7,577    & 1,632       & 4,474     & 4,990    & \multicolumn{1}{c|}{7,519}    & 34,806    \\
\multicolumn{1}{l|}{referrings}     & 52,513   & 22,496   & 2,911        & 6,808     & 4,990    & \multicolumn{1}{c|}{13,310}    & 103,028   \\
\multicolumn{1}{l|}{avg. boxes/ref} & 2.9      & 1.9      & 3.1         & 3.0       & 1.0       & \multicolumn{1}{c|}{0}         & 2.2    \\ \hline
\multicolumn{8}{c}{HumanRef Benchmark}                                                                                                  \\ \hline
\multicolumn{1}{l|}{type}           & attribute & position & interaction & reasoning & celebrity & \multicolumn{1}{c|}{rejection} & total \\
\multicolumn{1}{l|}{images}         & 838      & 972      & 940         & 982       & 1,000      & \multicolumn{1}{c|}{1,000}      & 5,732  \\
\multicolumn{1}{l|}{referrings}     & 1,000     & 1,000     & 1,000        & 1,000      & 1,000      & \multicolumn{1}{c|}{1,000}      & 6,000  \\
\multicolumn{1}{l|}{avg. boxes/ref}  & 2.8      & 2.1      & 2.1         & 2.7       & 1.1       & \multicolumn{1}{c|}{0}         & 2.2   \\ \hline
\end{tabular}}
\caption{
Main statistics of the HumanRef dataset, including the number of images, the number of referring expressions, the average word count per referring expression, and the average number of instances associated with each referring expression.}
\label{tab:sta_human}
\vspace{-2mm}
\end{table}

\begin{table}[]
  \centering
  \resizebox{1\columnwidth}{!}{%
   \begin{tabular}{cccccccc}
\hline
Datasets  & images & refs   & vocabs & \begin{tabular}[c]{@{}c@{}}avg. \\ size\end{tabular} & \begin{tabular}[c]{@{}c@{}}avg. \\ person/image\end{tabular} & \begin{tabular}[c]{@{}c@{}}avg. \\ words/ref\end{tabular} & \begin{tabular}[c]{@{}c@{}}avg. \\ boxes/ref\end{tabular} \\ \hline
RefCOCO~\cite{Datasets:REFCOCO}   & 1,519  & 10,771 & 1,874  & 593x484                                              & 5.72                                                          & 3.43                                                      & 1                                                         \\
RefCOCO+~\cite{Datasets:REFCOCO}  & 1,519  & 10,908 & 2,288  & 592x484                                              & 5.72                                                          & 3.34                                                      & 1                                                         \\
RefCOCOg~\cite{Datasets:REFCOCOG}  & 1,521  & 5,253  & 2,479  & 585x480                                              & 2.73                                                          & 9.07                                                      & 1                                                         \\ \hline
HumanRef & 5,732  & 6,000   & 2,714  & 1432x1074                                            & 8.60                                                          & 6.69                                                      & 2.2                                                       \\ \hline
\end{tabular}}
\caption{
Comparison of the HumanRef Benchmark with RefCOCO/+/g. For a fair comparison, we present only the statistics related to human referring in RefCOCO/+/g.}
\label{tab:sta_refcoco}
\vspace{-2mm}
\end{table}

\begin{figure}[t]
\centering
\includegraphics[width=1.0\linewidth]{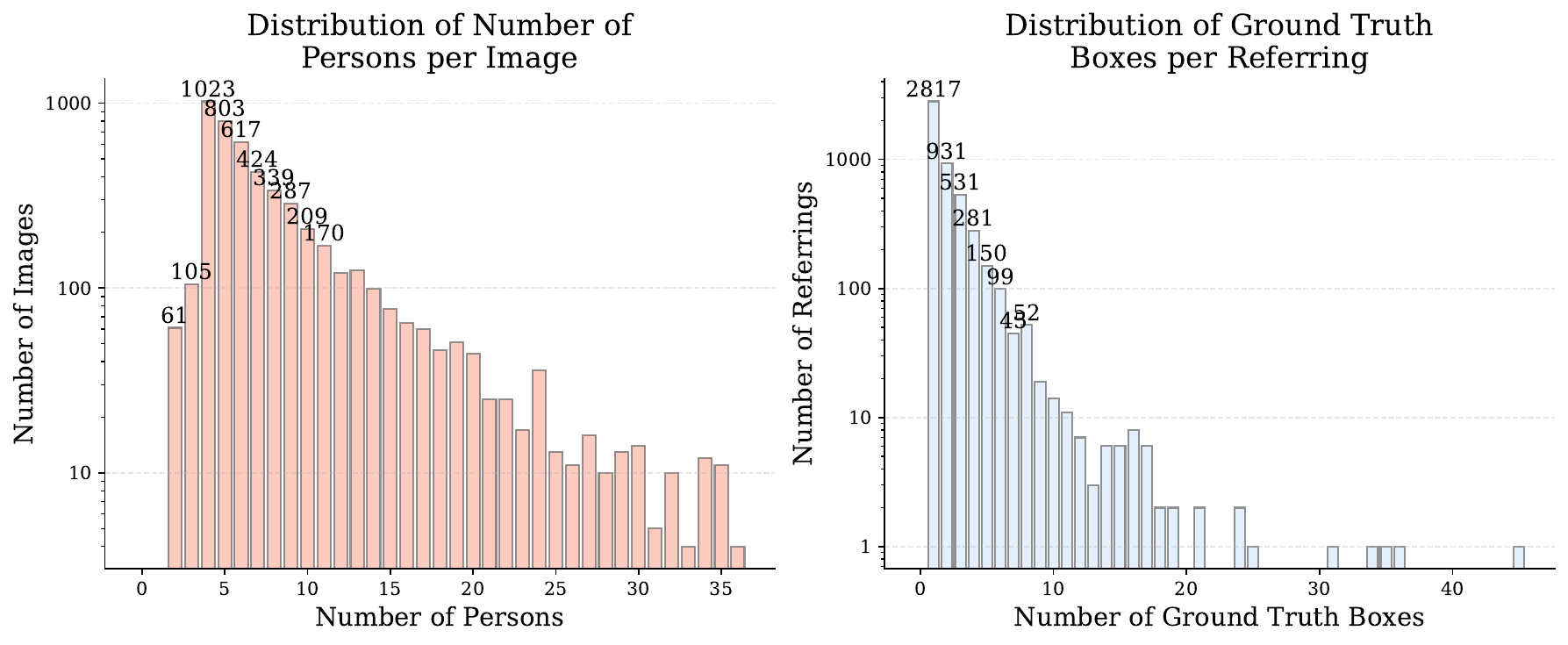}\vspace{-1mm}
\caption{Distribution of the number of individuals per image and the number of individuals referenced by each referring expression. }
\label{fig:statistics} 
\vspace{-3mm}
\end{figure}

\subsection{Statistics}
We first present the basic statistics of the HumanRef dataset and its subsets in Table \ref{tab:sta_human}, and then illustrate the characteristics of multi-instance referring and multi-instance discrimination in HumanRef in Figure \ref{fig:statistics}. Additionally, Table \ref{tab:sta_refcoco} compares the HumanRef Benchmark with widely used referring benchmarks, including RefCOCO, RefCOCO+, and RefCOCOg. A key distinction of HumanRef is its higher image resolution and larger number of individuals per image, requiring models to precisely identify all correct individuals among multiple people. Unlike traditional benchmarks, where each referring expression corresponds to a single person, HumanRef supports multi-instance referring, offering a more realistic and challenging evaluation setting for referring expression comprehension.
\section{RexSeek Model}
\label{seuc:method}

\subsection{Model Design Philosophy}
From a model design perspective, we argue that a robust referring model should have two essential capabilities: \textbf{i) robust perception ability}, where the model can reliably detect all individuals in an image, and \textbf{ii) strong language comprehension}, where the model can accurately interpret complex natural language descriptions of people.

For the first capability, modern object detection models~\cite{jiang2025t, ren2024dino, liu2023grounding, ren2024grounding} are highly effective at identifying people within images. However, these models often lack the necessary language comprehension abilities to process intricate and nuanced referring expressions. On the other hand, while MLLMs are proficient in understanding natural language, they often struggle with fine-grained object detection tasks. Inspired by ChatRex~\cite{jiang2024chatrex}, we propose a hybrid framework, RexSeek, which integrates the strengths of both object detection models and LLMs. RexSeek combines a high-performance detection model with a multimodal LLM to achieve both accurate detection and effective language understanding.

\begin{figure}[t]
\centering
\includegraphics[width=1.0\linewidth]{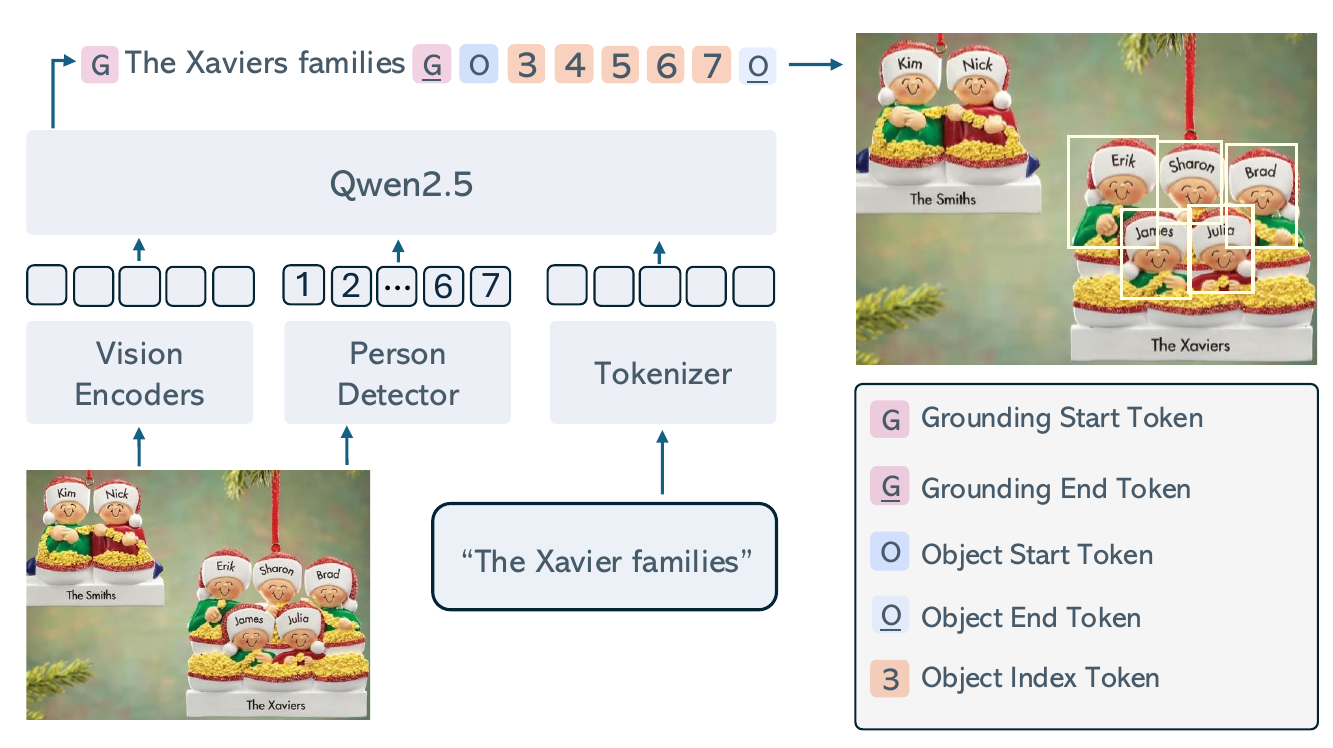}\vspace{-1mm}
\caption{Overview of the RexSeek model. RexSeek is a retrieval-based model built upon ChatRex~\cite{jiang2024chatrex}. By integrating a person detection model, RexSeek transforms the referring task from predicting box coordinates to retrieving the index of input boxes.}
\label{fig:rexseek} 
\vspace{-3mm}
\end{figure}

\begin{table*}[t]
\centering
  \resizebox{1.0\linewidth}{!}{
    \begin{tabular}{c|ccc|ccc|ccc|ccc|ccc|ccc|c}
\toprule
\multirow{2}{*}{Method}                     & \multicolumn{3}{c|}{Attribute} & \multicolumn{3}{c|}{Position} & \multicolumn{3}{c|}{Interaction} & \multicolumn{3}{c|}{Reasoning} & \multicolumn{3}{c|}{Celebrity} & \multicolumn{3}{c|}{Average} & \multicolumn{1}{l}{Rejection} \\ \cline{2-20} 
                                            & R         & P       & DF1     & R         & P       & DF1     & R          & P        & DF1      & R         & P        & DF1     & R         & P        & DF1     & R        & P       & DF1     & Score                         \\ \midrule
\multicolumn{1}{c|}{Baseline$\dagger$} & 100.0     & 37.2    & 24.2    & 100.0     & 28.5    & 15.9    & 100.0      & 32.5     & 19.4     & 100.0     & 42.6     & 30.3    & 100.0     & 14.4     & 4.9     & 100.0    & 31.0    & 18.9    & 0.0                             \\
DINOX~\cite{ren2024dino}                                       & 59.5      & 28.8    & 20.9    & 78.8      & 28.1    & 17.6    & 67.3       & 28.5     & 18.9     & 76.2      & 32.1     & 22.2    & 94.1      & 48.0     & 37.0    & 75.2     & 33.1    & 23.3    & 36.0                            \\
InternVL-2.5-8B~\cite{chen2024expanding}                             & 23.5      & 39.0    & 27.1    & 23.0      & 28.0    & 24.3    & 27.8       & 40.1     & 31.3     & 17.5      & 22.8     & 18.9    & 57.4      & 59.3     & 58.0    & 29.8     & 37.8    & 31.9    & 54.9                           \\
Ferret-7B~\cite{you2023ferret}                                   & 27.9      & 44.4    & 30.4    & 30.2      & 36.2    & 29.8    & 30.8       & 41.8     & 31.2     & 19.7      & 33.7     & 22.8    & 63.2      & 60.0     & 57.5    & 34.4     & 43.2    & 34.3    & 2.0                             \\
Groma-7B~\cite{ma2024groma}                                    & 67.5      & 47.8    & 38.6    & 63.2      & 43.1    & 37.2    & 66.6       & 48.1     & 40.6     & 59.1      & 41.4     & 34.8    & 73.2      & 63.3     & 59.1    & 65.9     & 48.7    & 42.1    & 0.0                             \\
ChatRex-7B~\cite{jiang2024chatrex}                                  & 44.3      & 78.0    & 51.8    & 48.0      & 66.7    & 52.5    & 49.6       & 74.8     & 56.5     & 36.6      & 65.1     & 42.8    & 73.7      & 76.5     & 74.2    & 50.4     & 72.2    & 55.6    & 0.0                             \\
Qwen2.5-VL-7B~\cite{bai2025qwen2}                               & 49.1      & 71.3    & 54.4    & 50.2      & 61.7    & 52.8    & 48.2       & 66.3     & 53.2     & 34.6      & 61.2     & 40.3    & 80.3      & 81.9     & 80.1    & 52.5     & 68.5    & 56.2    & 7.1                            \\
DeepSeek-VL2-small~\cite{wu2024deepseek}                          & 52.3      & 78.0    & 57.7    & 56.4      & 66.1    & 58.1    & 55.4       & 75.7     & 60.7     & 46.6      & 61.7     & 50.1    & 85.9      & 74.3     & 70.7    & 59.3     & 71.2    & 59.5    & 3.1                            \\
Molmo-7B-D$^*$~\cite{deitke2024molmo}                                  & 82.7      & 86.4    & 76.3    & 78.0      & 80.6    & 72.4    & 69.9       & 77.7     & 66.1     & 72.1      & 80.4     & 65.5    & 85.9      & 87.5     & 82.9    & 77.7     & 82.5    & 72.6    & 68.6                           \\ \midrule
\rowcolor{gray!15}RexSeek-7B                                  & 87.2      & 86.8    & 81.5    & 86.1      & 86.3    & 83.8    & 84.8       & 84.6     & 80.7     & 87.8      & 84.7     & 81.5    & 83.4      & 86.5     & 84.2    & 85.9     & 85.8    & 82.3    & 54.1                           \\ \bottomrule
\end{tabular}}
  \caption{Benchmarking multimodal models on HumanRef Benchmark. R, P, and DF1 represent Recall, Precision, and DensityF1, respectively. $\dagger$ A simple baseline that uses the bounding boxes of all persons in the image as results, simulating a person detection model that does not follow the referring description. $^*$Molmo-7B-D predicts point coordinates as output and use point-in-mask evaluation criteria. }
  \label{tab:HumanRef_benchmark}
  \vspace{-2mm}
  \end{table*} 

\subsection{Architecture}
Following ChatRex, we formulate the referring task as a retrieval-based process~\cite{ma2024groma, jiang2024chatrex}. As illustrated in Figure \ref{fig:rexseek}, RexSeek consists of three main components: vision encoders, a person detector, and a large language model. Given an input image, we first pass it through a dual vision encoder module used in ChatRex. This module consists of a CLIP~\cite{VLP:CLIP} to extract low-resolution image features $\mathcal{F}_{\text{low}}$ and a ConvNeXt~\cite{liu2022convnet} to extract high-resolution image features $\mathcal{F}_{\text{high}}$. 
We adjust the input resolutions for both vision encoders to ensure they generate the same number of tokens at the last scale. The final vision tokens $\mathcal{F}$ is obtained by concatenating these features at the channel dimension:
\[
\mathcal{F} = \text{Concat}(\mathcal{F}_{\text{low}}, \mathcal{F}_{\text{high}})
\]
Next, we prompt DINO-X~\cite{ren2024dino} to get the bounding boxes of persons \( \{B_i\}_{i=1}^K \) in the image. For each bounding box, we extract its RoI features \( \mathcal{C}_i \) and add their positional embeddings to generate object tokens $\mathcal{O}_i$, which capture both the content and spatial context of each detected person:
\[
\mathcal{O}_i = \mathcal{C}_i + \operatorname{PE}(B_i)
\]

Specifically, the RoI feature is extracted from the high-resolution vision features using a multi-scale RoI Align operation~\cite{he2017mask}. The positional embedding is computed by encoding the bounding box coordinates $(x, y, w, h)$ using a sinusoidal encoding function and concatenating the encoded values along the channel dimension.

Finally, the vision tokens $\mathcal{F}$, object tokens \( \mathcal{O}\), and text tokens \( \mathcal{T} \) are projected using different MLPs and then fed into the LLM. By default, we use Qwen2.5~\cite{yang2024qwen2} as the LLM. The LLM decodes the input to produce the corresponding box indices \( \mathcal{I} \):
\[
\mathcal{I} = \text{LLM}(\mathcal{F}, \mathcal{O}, \mathcal{T})
\]
The output \( \mathcal{I} \) consists of object indices that correspond to the bounding boxes of the target persons corresponding to the referring. This sequence is structured as follows:
\[
\small
\texttt{<g>referring</g><o><objm>...<objn></o>}
\]
Here, \(\texttt{<objm>}\) and \(\texttt{<objn>}\) refer to specific object index tokens that correspond to the detected persons. The special tokens \(\texttt{<g>}\), \(\texttt{</g>}\), \(\texttt{<o>}\), and \(\texttt{</o>}\) are used to format the output, linking the referring expression with the relevant object indices.

\begin{table}[]
  \centering
  \resizebox{1\columnwidth}{!}{%
   \begin{tabular}{c|lcc|c}
\toprule
Stage  & \begin{tabular}[c]{@{}l@{}}Trainable\\ Modules\end{tabular}                                & Task                                                                                                & \# Samples & Datasets                                                                \\ \midrule
Stage1 & \multicolumn{1}{c}{MLPs}                                                                   & Image Captioning                                                                                    & 976K       & ALLAVA-4V-Caption~\cite{chen2024allava}                                                       \\ \midrule
Stage2 & \multicolumn{1}{c}{\begin{tabular}[c]{@{}c@{}}MLPs + LLM +\\ Vision Encoders\end{tabular}} & \begin{tabular}[c]{@{}c@{}}Grounding \& \\ Region Understanding\end{tabular}                        & 2.07M      & \begin{tabular}[c]{@{}c@{}}COCO~\cite{Datasets:MSCOCO}, LVIS~\cite{gupta2019lvis},\\ O365~\cite{shao2019objects365}, Rexverse-2M~\cite{jiang2024chatrex}\end{tabular} \\ \hline
Stage3 & \begin{tabular}[c]{@{}l@{}}MLPs + LLM + \\ Vision Encoders\end{tabular}                    & \begin{tabular}[c]{@{}c@{}}General Knowledge  \&\\ Grounding \&\\ Region Understanding\end{tabular} & 2.15M      & \begin{tabular}[c]{@{}c@{}}LLAVA-665K~\cite{VLM:LLaVA-1.5} \\ Rexverse-2M~\cite{jiang2024chatrex}\end{tabular}        \\ \midrule
Stage4 & \begin{tabular}[c]{@{}l@{}}MLPs + LLM +\\ Vision Encoders\end{tabular}                     & Referring                                                                                           & 103K        & HumanRef                                                               \\ \bottomrule
\end{tabular}}
\caption{
Data, task, and trainable modules for each stage.}
\label{tab:training}
\vspace{-2mm}
\end{table}

\subsection{Four Stage Training}
Similar to other VLMs, we adopt a pretraining followed by supervised fine-tuning approach~\cite{VLM:LLaVA}. Our training process consists of four stages. In the first stage, we align the visual and textual modalities using image-captioning data. In the second stage, we focus on perception training with detection-oriented data, enabling the model to retrieve relevant objects from input bounding boxes. In the third stage, we incorporate multimodal data to enhance the model’s general understanding abilities. Finally, in the fourth stage, we fine-tune the model using the HumanRef dataset, resulting in the final RexSeek model. The data, task, and trainable modules for each stage are shown in Table \ref{tab:training}.

\section{Experiments}
\label{seuc:experiments}
In this section, we first introduce the evaluation metrics used in our study and assess the performance of multimodal models on HumanRef. We perform a comprehensive analysis to explore the challenges faced by existing models in handling the referring task. Additionally, we perform ablation experiments on RexSeek for model design choices.

\subsection{Metrics}
We evaluate the referring task using Precision, Recall, and DensityF1 Score. Given a referring expression, the model predicts one or more bounding boxes, and a prediction is considered correct if its IoU with any ground truth box exceeds a predefined threshold. Following the evaluation protocol in COCO~\cite{Datasets:MSCOCO}, we report the average performance across IoU thresholds from 0.5 to 0.95 in increments of 0.05. For models that only output points, such as Molmo~\cite{deitke2024molmo}, a prediction is considered correct if the predicted point falls within the mask of the corresponding instance.  However, this evaluation is less strict than the IoU-based metric, as point-in-mask criteria impose looser spatial constraints, making direct comparisons less fair. For the rejection subset, we calculate the number of referring expressions that the model does not predict any boxes and divide it by the number of total expressions.

To penalize models that indiscriminately detect all persons in an image to achieve a high F1 score through high recall, we introduce the DensityF1 Score, which modifies the standard F1 Score with a density-aware penalty:
\begin{equation}
\text{DensityF1} = \frac{1}{N} \sum_{i=1}^{N} 2 \times \frac{\text{Precision}_i \times \text{Recall}_i}{\text{Precision}_i + \text{Recall}_i} \times D_i
\end{equation}
where \( D_i \) is the density penalty factor, defined as:
\begin{equation}
D_i = \min(1.0, \frac{\text{GT Count}_i}{\text{Predicted Count}_i})
\end{equation}

Here, GT Count is the total number of persons in an image, and Predicted Count is the number of predicted boxes for a given referring expression. This penalty discourages over-detection by reducing the score when the predicted box count significantly exceeds the ground truth count.

\begin{figure}[t]
\centering
\includegraphics[width=1.0\linewidth]{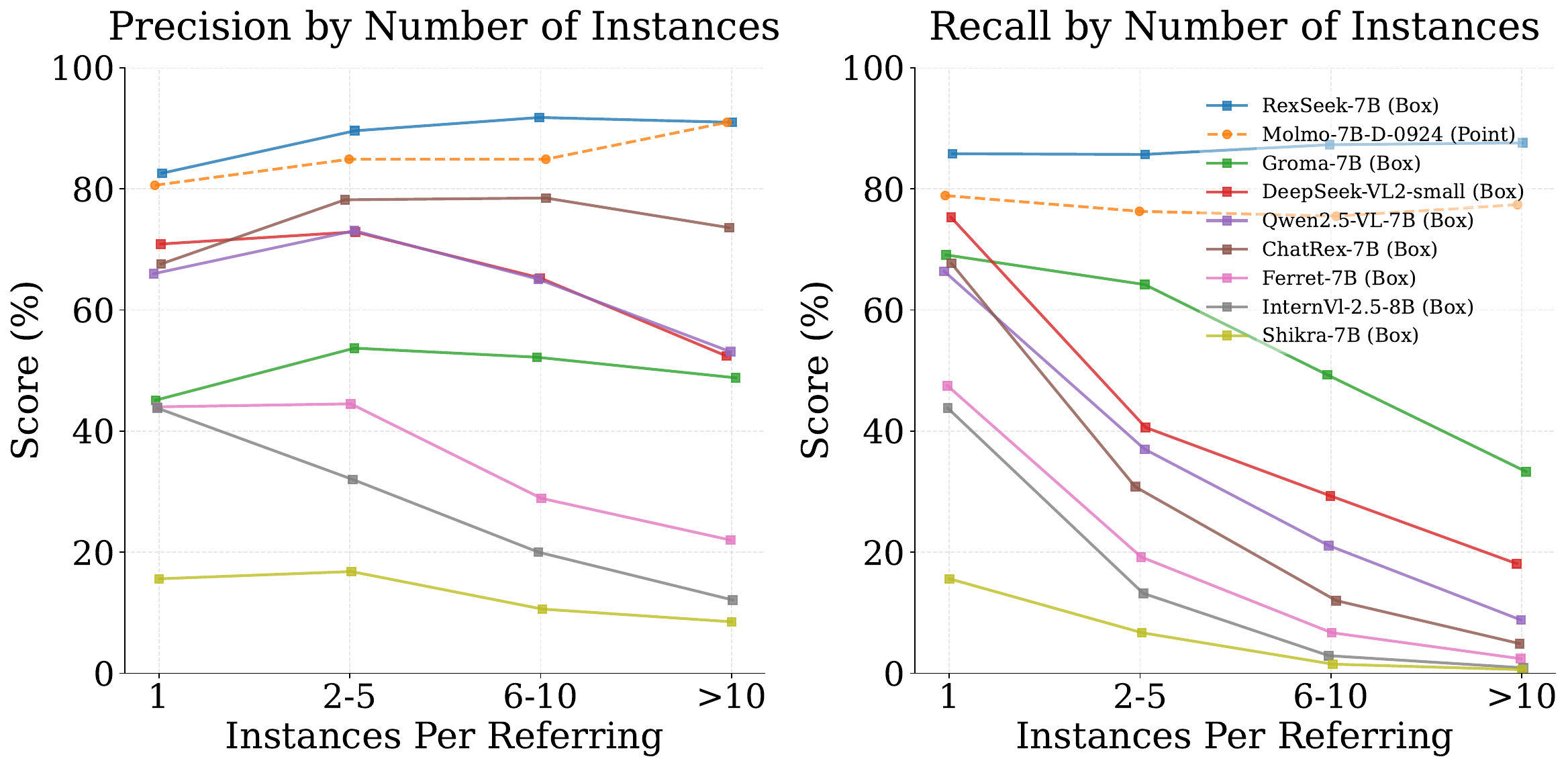}\vspace{-1mm}
\caption{Visualizing the trend of recall and precision variations across different models as the number of instances corresponding to each referring expression increases.}
\label{fig:rp_box_count} 
\vspace{-2mm}
\end{figure}

\subsection{Benchmarking on HumanRef}
In Table \ref{tab:HumanRef_benchmark}, we evaluate the performance of various multimodal models on the HumanRef benchmark. While these models perform well on the widely used RefCOCO, RefCOCO+, and RefCOCOg benchmarks, their performance significantly degrades on HumanRef. Our analysis reveals two common issues among these models:

\textbf{Low Recall for Multi Instance:}
We observe a common issue among most models: when a referring expression corresponds to multiple instances, recall drops significantly, as shown in Figure \ref{fig:rp_box_count}. This suggests that when multiple objects need to be detected, most models tend to predict only a few bounding boxes, limiting their applicability in real-world scenarios. A key factor contributing to this behavior is the nature of the training data. Most multimodal models are trained on RefCOCO, RefCOCO+, and RefCOCOg, where referring expressions rarely correspond to multiple instances. As a result, these models become biased toward single-instance predictions.
In contrast, RexSeek has been trained on datasets that explicitly include multi-instance referring expressions, demonstrate a significantly improved ability to handle these real-world cases.

\textbf{Hallucination Issue:}
On the rejection subset, we observe that most models perform poorly with low rejection score. This indicates that regardless of whether the referred object is actually present in the image, these models tend to predict a bounding box, exhibiting a severe hallucination issue. In real-world referring applications, such as referring in video streams, it is crucial for models to accurately determine whether the specified object exists in the image. Additionally, we find that the rejection capability can be significantly improved by incorporating appropriate training data. As shown in Table \ref{tab:rejection}, when trained without the rejection data in HumanRef, RexSeek also demonstrates strong hallucination tendencies. This highlights the critical role of dataset design in the referring task, as inadequate dataset construction can lead to overconfident predictions.

\begin{table}[]
  \centering
  \resizebox{0.7\columnwidth}{!}{%
   \begin{tabular}{c|cc}
\toprule
Model      & With Rejection Data & Rejction Score \\ \midrule
RexSeek-7B & No                      &     0           \\
RexSeek-7B & Yes                     &     541         \\ \bottomrule
\end{tabular}}
\caption{
Rejection score comparison under different model scales with and without rejection data during training.}
\label{tab:rejection}
\vspace{-2mm}
\end{table}

\begin{table}[]
  \centering
  \resizebox{0.55\columnwidth}{!}{%
   \begin{tabular}{c|ccc}
\toprule
\multirow{2}{*}{Loading Stage} & \multicolumn{3}{c}{HumanRef Average} \\
                               & R           & P          & DF1        \\ \midrule
stage1                         & 73.9        & 73.5       & 68.2       \\
stage2                         & 77.0        & 77.3       & 72.2       \\
stage3                         & 77.9        & 78.0       & 73.0       \\ \bottomrule
\end{tabular}}
\caption{
Ablation experiments on multi-stage training by loading models from different training stages and fine-tuning them on the HumanRef dataset. We Qwen2.5-3B as the base LLM.}
\label{tab:ab1}
\vspace{-2mm}
\end{table}

\begin{table}[]
  \centering
  \resizebox{0.5\columnwidth}{!}{%
   \begin{tabular}{c|cc}
\hline
\multirow{2}{*}{Method} & \multicolumn{2}{c}{RefCOCOg} \\ \cline{2-3} 
                        & val           & test         \\ \hline
Shikra-7B~\cite{chen2023shikra}               & 82.3          & 82.2         \\
InternVL2-8B~\cite{chen2024expanding}            & 82.7          & 82.7         \\
Grounding DINO-L~\cite{liu2024grounding}        & 86.1          & 87.0         \\
Qwen2.5-VL-7B~\cite{bai2025qwen2}           & 87.2          & 87.2         \\
MM1.5-7B~\cite{zhang2024mm1}                & -             & 87.1         \\
ChatRex-7B~\cite{jiang2024chatrex}              & 88.8          & 88.6         \\ \hline
RexSeek-7B              & 84.0          & 84.4         \\ \hline
\end{tabular}}
\caption{
Zero-shot evaluation of RexSeek on RefCOCOg. We use the open-set detector DINOX to detect the subject object in the image and use the detected bounding box as input to RexSeek.}
\label{tab:refcoco_ab}
\vspace{-3mm}
\end{table}

\subsection{Ablations on RexSeek}
\textbf{Ablation of Multi-stage Training:}
We analyzed the impact of the four-stage training approach used in RexSeek. As shown in Table \ref{tab:ab1}, we conducted supervised fine-tuning on the HumanRef dataset after each training stage. The results demonstrate that the model achieves its best performance after undergoing SFT with general multimodal data (LLaVA-665K~\cite{VLM:LLaVA-1.5}). We attribute this improvement to the model acquiring richer general knowledge from multimodal data, which enhances its ability to accurately refer to persons in complex scenarios.

\begin{figure}[t]
\centering
\includegraphics[width=0.9\linewidth]{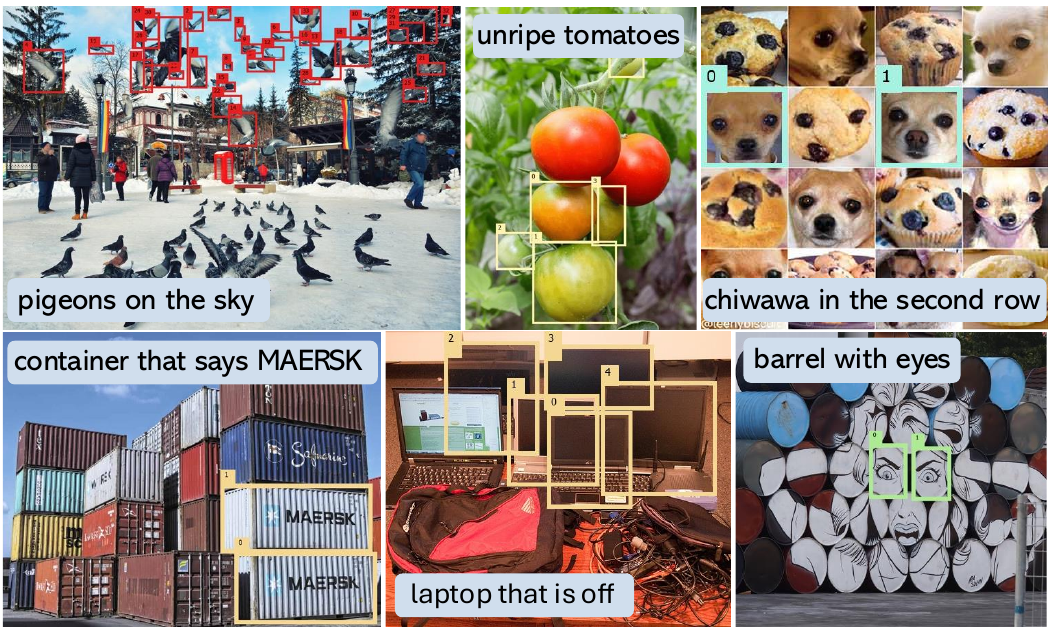}\vspace{-1mm}
\caption{RexSeek can refer to arbitrary objects beyond person.}
\label{fig:anyobject} 
\vspace{-3mm}
\end{figure}

\textbf{Generalization to Any Object Referring:}
Although RexSeek is trained exclusively on human-related referring data, we find that it also demonstrates the ability to refer to arbitrary objects. We first evaluate the performance of RexSeek on RefCOCOg. Given a referring expressions, we apply DINO-X to detect the main object in the image, using the detected bounding box as input to RexSeek. As shown in Table \ref{tab:refcoco_ab}, RexSeek achieves competitive performance on RefCOCO/+/g, despite not being explicitly trained on general object referring. Additionally, Figure \ref{fig:anyobject} presents visualizations illustrating that RexSeek can also detect multiple instances even for non-human objects. We attribute this generalization ability to our multi-stage training approach, where perception and multimodal understanding training develop object comprehension, and fine-tuning on HumanRef effectively extends it to arbitrary objects.

\section{Conclusion}
\label{seuc:conclusion}
In this work, we identify the fundamental limitations of existing referring datasets and models, demonstrating that they fail to meet real-world application demands, particularly in multi-instance referring. To address this, we introduce HumanRef, a large-scale benchmark reflecting real-world complexity, and propose RexSeek, a retrieval-based detection MLLM integrating person detection with a language model. Our multi-stage training approach equips RexSeek with strong generalization capabilities, allowing it to excel in human-centric referring while extending effectively to arbitrary object referring. Extensive evaluations highlight the struggles of state-of-the-art models with multi-instance detection and hallucination, underscoring the importance of dataset design and training strategies for more reliable and generalizable referring expression models.
{
    \small
    \bibliographystyle{ieeenat_fullname}
    \bibliography{main}
}

\clearpage
\setcounter{page}{1}
\maketitlesupplementary

\section{More Details of HumanRef Dataset}
\subsection{Detailed Definition for Each Subset}
\textbf{Attribute Subset:}
The attribute subset in HumanRef encompasses a diverse range of descriptive properties used for referring to individuals. These attributes are systematically categorized into finite-value attributes, which have a predefined set of possible values, and open-ended attributes, which allow for more flexible and detailed descriptions. As illustrated in Fig. \ref{lst:humanref_attributes}, we provide a structured breakdown of these attribute categories along with representative examples. This visualization highlights the key properties that can be used to distinguish individuals, including intrinsic characteristics (e.g., gender, age, ethnicity), appearance features (e.g., hairstyle, facial expressions), clothing and accessories, poses, and actions. By incorporating both structured and descriptive attributes, HumanRef ensures a comprehensive and versatile annotation framework that better aligns with real-world referring scenarios.

\noindent\textbf{Position Subset:}
The position subset in HumanRef captures the spatial relationships of individuals within an image. We categorize these into two main types: inner position and outer position, which represent different spatial referencing strategies. Inner position describes the relative spatial relationships between individuals, while outer position refers to absolute spatial relationships using environmental landmarks. We show some detailed examples in Table. \ref{tab:position_examples}.

\begin{table}[h]
    \centering
    \small
    \renewcommand{\arraystretch}{1.2} 
    \begin{tabular}{p{0.45\linewidth} | p{0.45\linewidth}}
        \toprule
        \textbf{Inner Position (Relative to Others)} & \textbf{Outer Position (Relative to Environment)} \\
        \midrule
        \textit{``The leftmost person in the image.''} & \textit{``The person standing in the corner of the room.''} \\
        \textit{``The second person from the left.''} & \textit{``The person sitting on the red chair.''} \\
        \textit{``The person at the top of the group.''} & \textit{``The person standing at the left edge of the bridge.''} \\
        & \textit{``The person leaning against the wall next to a tree.''} \\
        & \textit{``The person standing near the window of the room.''} \\
        \bottomrule
    \end{tabular}
    \caption{Examples of Inner and Outer Position References.}
    \label{tab:position_examples}
\end{table}

\lstdefinelanguage{json}{
    basicstyle=\ttfamily\footnotesize, 
    numbers=left,
    numberstyle=\scriptsize,
    stepnumber=1,
    showstringspaces=false,
    breaklines=true,
    frame=lines,
    backgroundcolor=\color{gray!10},
    morestring=[b]",
    morecomment=[l]{//},
    commentstyle=\color{gray},
    stringstyle=\color{teal}
}

\begin{figure*}[t] 
    \centering    
    \begin{lstlisting}[language=json]
{
    "Gender": ["Male", "Female"],
    "Age": ["Infant", "Child", "Adolescent", "Adult", "Elderly"],
    "Ethnicity": ["White", "African", "Asian", "..."],
    "Profession": ["Doctor", "Engineer", "Teacher", "..."],
    "Appearance": {
        "Hair": {
            "Type": ["Short", "Long", "Curly", "Bald", "..."],
            "Color": ["Black", "Brown", "Blonde", "Red", "..."]
        },
        "Beard": {
            "Type": ["Full Beard", "Goatee", "Mustache", "..."],
            "Color": ["Black", "White", "Brown", "..."]
            "Detailed": ["long and black beard", "..."]
        },
        "Facial Expression": {
            "Type": ["Smiling", "Frowning", "Surprised", "..."],
            "Detailed": ["Smiling with closed eyes", "..."]
        }
    },
    "Clothing & Accessories": {
        "Upper Body": {"Type": ["T-shirt", "Shirt", "..."], "Color": ["Red", "Blue", "..."]},
        "Lower Body": {"Type": ["Jeans", "Skirt", "..."], "Color": ["Black", "White", "..."]},
        "Shoes": {"Type": ["Sneakers", "Boots", "..."], "Color": ["Red", "Blue", "..."]},
        "Hat": {"Type": ["Baseball Cap", "Knit Cap", "..."], "Color": ["Black", "White", "..."]}
    },
    "Pose": {"Type": ["Standing", "Sitting", "..."], "Details": ["Sitting on a bench", "..."]},
    "Actions": {"Type": ["Walking", "Running", "..."], "Details": ["Walking a dog", "..."]}
}
    \end{lstlisting}
    \caption{Attribute taxonomy in HumanRef. Finite-value attributes have predefined values, while open-ended attributes allow flexible descriptions.}
    \label{lst:humanref_attributes}
\end{figure*} 

\noindent\textbf{Interaction Subset:}
The interaction subset in HumanRef captures the ways individuals interact with other people and objects within a scene. We classify interactions into two categories: inner interaction and outer interaction. Inner interaction focuses on actions between individuals, while outer interaction describes interactions between a person and objects or the surrounding environment. We show some detailed examples in Table. \ref{tab:interaction_examples}.

\begin{table}[h]
    \centering
    \small
    \renewcommand{\arraystretch}{1.2} 
    \begin{tabular}{p{0.45\linewidth} | p{0.45\linewidth}}
        \toprule
        \textbf{Inner Interaction (Human-Human)} & \textbf{Outer Interaction (Human-Object/Environment)} \\
        \midrule
        \textit{``Two players making physical contact.''} & \textit{``The person holding a dog on a leash.''} \\
        \textit{``The bride and groom walking hand in hand in the middle.''} & \textit{``The person reaching out to grab the football.''} \\
        \textit{``The person raising another person's hand with their right hand.''} & \textit{``The person using their right hand to pull toilet paper.''} \\
        \textit{``Two people embracing each other.''} & \textit{``The person holding a transparent box containing a roll of toilet paper with their left hand.''} \\
        & \textit{``The person holding a pizza with one hand.''} \\
        & \textit{``The person holding a pen in their hand.''} \\
        \bottomrule
    \end{tabular}
    \caption{Examples of Inner and Outer Interaction References.}
    \label{tab:interaction_examples}
\end{table}

\noindent\textbf{Reasoning Subset:} 
The reasoning subset in HumanRef requires models to perform multi-step inference by first identifying a reference person or object before determining the target individual. It is divided into three categories: inner position reasoning, outer position reasoning, and attribute reasoning. Inner position reasoning describes one individual using another as an anchor point, establishing a relative spatial relationship between them. This approach ensures that the model must first locate a reference person before resolving the intended target. Outer position reasoning follows a similar principle but instead uses an absolute spatial reference tied to the surrounding environment, requiring the model to integrate positional understanding of both individuals and external scene elements. 

Attribute reasoning involves logical filtering within a group of candidates who share a common attribute, followed by an exclusion step based on an additional attribute with a negation rule. This forces the model to refine its selection beyond direct attribute matching, ensuring a more precise understanding of distinguishing characteristics. These three types of reasoning introduce hierarchical complexity into the referring task, strengthening the model’s ability to process contextual, spatial, and attribute-based relationships in a structured manner. We show some detailed examples in Table. \ref{tab:reasoning_examples}.

\begin{table}[h]
    \centering
    \small
    \renewcommand{\arraystretch}{1.2} 
    \begin{tabular}{p{0.3\linewidth} | p{0.6\linewidth}}
        \toprule
        \textbf{Reasoning Type} & \textbf{Example Expressions} \\
        \midrule
        \textbf{Inner Position Reasoning} & \textit{``The person to the left of the child wearing a blue-and-white striped shirt.''} \\
        & \textit{``The person to the right of the man in a suit.''} \\
        & \textit{``The girl to the right of the person wearing blue headphones.''} \\
        & \textit{``All individuals to the right of the groom.''} \\
        \midrule
        \textbf{Outer Position Reasoning} & \textit{``The closest person to the left of the person in the corridor.''} \\
        & \textit{``The person to the left of the individual directly below the letter D.''} \\
        & \textit{``The child to the right of the girl standing under the blue arched wooden door.''} \\
        & \textit{``The person to the right of the individual sitting inside the shopping cart.''} \\
        \midrule
        \textbf{Attribute Reasoning} & \textit{``The person wearing a hat but not sitting.''} \\
        & \textit{``The person eating ice cream but not wearing a red top.''} \\
        & \textit{``The person wearing sandals but not sitting on the bed.''} \\
        & \textit{``The person on the airplane but not showing their head.''} \\
        \bottomrule
    \end{tabular}
    \caption{Examples of Reasoning-Based Referring Expressions.}
    \label{tab:reasoning_examples}
\end{table}

\noindent\textbf{Celebrity Subset:} 
The celebrity subset in HumanRef focuses on identifying well-known individuals based on their names or recognizable personas. To provide a structured classification, we divide celebrities into six categories: Character, Singer, Actor, Athlete, Entrepreneur, and Politician. This categorization ensures that the dataset covers a diverse range of public figures from different domains, each of whom may be referred to by their real name or an associated identity.  In Tab. \ref{tab:celebrity}, we provide a detailed list of representative individuals from each category, illustrating the range of celebrities included in the dataset. 

\subsection{Details for Structured Property Dictionary}
 To facilitate the annotation process, we employ Qwen2.5-VL-7B to generate a predefined structured property dictionary for each person in the image. This dictionary serves as a reference for annotators, allowing them to construct property lists based on pre-generated attribute descriptions. Figure \ref{fig:structured_prompt} presents the prompt used to generate these structured descriptions.

\section{Details for RexSeek model}
\subsection{Model Architecture}
We utilize a dual-encoder design for vision processing. The low-resolution visual encoder is based on the CLIP ViT-Large-14-336 model, while the high-resolution visual encoder leverages the LAION ConvNeXt-Large-320 model. The input resolution is set to 336×336 for the low-resolution encoder and 768×768 for the high-resolution encoder, allowing the model to capture both coarse and fine-grained visual details effectively.

\subsection{Training Details}
During the pretraining stage (Stage-1), we use a batch size of 32 per device, resulting in a total batch size of 256 across all devices. The instruction-tuning stage (Stage-2, Stage-3, Stage-4) employs a reduced batch size of 16 per device, with a total batch size of 128. The learning rate is initialized at 1e-3 for pretraining and adjusted to 2e-5 during instruction tuning to ensure stable fine-tuning and convergence.

\begin{table*}[t]
\centering
  \resizebox{1.0\linewidth}{!}{
    \begin{tabular}{l|l}
\hline
Character    & \begin{tabular}[c]{@{}l@{}}Eleven from the Strange Things, Obi-Wan Kenobi, Captain America, Queen Maeve, Buzz Lightyear, Mary Poppins, John Rambo, Hermione Granger, Rick Grimes, \\ Ferris Bueller, Sheldon Cooper, Sarah Connor, Negan Smith, Amy Farrah Fowler, Professor Severus Snape, Shrek, Ada Shelby, Palpatine, Jorah Mormont, Thomas Shelby, \\ Cersei Lannister, Luke Skywalker, Maximus Decimus Meridius, Sansa Stark, John Wick, Atticus Finch, Harry Potter, Wolverine, Mr. Bean, Ronald Weasley, \\ Melisandre, Ross Geller, Bilbo Baggins, Samwise Gamgee, Maggie Greene, Jon Snow, Wednesday Addams, Hans Gruber, Rocky Balboa, Michael Corleone, \\ Leonard Hofstadter, Chandler Bing, Littlefinger, Barbossa, Rachel Green, Howard Wolowitz, Jaime Lannister, Legolas, Axel Foley, James T. Kirk, Kevin McCallister, \\ Margaery Tyrell, Leia Organa, Forrest Gump, Marty McFly, Han Solo, Billy Butcher, Mr. Kesuke Miyagi, Professor Albus Dumbledore, Steve Harrington, Sirius Black, \\ Gus Fring, Jack Sparrow, Inigo Montoya, Professor Minerva McGonagall, Homelander, Arthur Shelby, Superman, Indiana Jones, Beetlejuice, Polly Gray, Hughie Campbell, \\ Peter Venkman, Sandor Clegane, Soldier Boy, Joey Tribbiani, Rajesh Koothrappali, Gollum, Vito Corleone, Daryl Dixon, Aragorn, Andy Dufresne, Jesse Pinkman, \\ Penny Hofstadter, Jean-Luc Picard, Lord Voldemort, Oberyn Martell, Luna Lovegood, Carol Peletier, Monica Geller, Alfred Pennyworth, Bernadette Rostenkowski-Wolowitz, \\ Darth Maul, Thor, Neo, John McClane, Phoebe Buffay, Spock, Dorothy Gale\end{tabular}                                                                                                                                                                                                                                                                                                                                                                                                                                                                                                                                                                                                                                                                                                                                                                                                                                                                                                                                                                                                                                                                                                                                                                                                                                                     \\ \hline
Singer       & \begin{tabular}[c]{@{}l@{}}Michael Bublé, Demi Lovato, Aerosmith, Drake, ZAYN, Jennifer Lopez, Olivia Rodrigo, Kali Uchis, Flo Rida, Doja Cat, Skrillex, Chris Brown, Ellie Goulding, 50 Cent, \\ Katy Perry, Gucci Mane, Charli XCX, Avril Lavigne, Shawn Mendes, Lil Wayne, J. Cole, Linkin Park, Migos, Taylor Swift, DJ Khaled, Red Hot Chili Peppers, Justin Bieber, \\ Calvin Harris, 2 Chainz, Frank Ocean, Jason Mraz, Alicia Keys, Miley Cyrus, Childish Gambino, Meghan Trainor, Tyga, Usher, SZA, Bad Bunny, Labrinth, Diplo, \\ Jack Johnson, Halsey, Young Thug, Bon Jovi, Post Malone, Christina Aguilera, The Kooks, John Mayer, The 1975, Akon, G-Eazy, Panic! At the Disco, Eminem, Ed Sheeran,\\  Maroon 5, Ne-Yo, Zedd, Dr. Dre, Queen, Nelly Furtado, Steve Lacy, Imagine Dragons, Sia, Mac DeMarco, Big Sean, Martin Garrix, Camila Cabello, The Rolling Stones, \\ Khalid, Harry Styles, Charlie Puth, Kanye West, The Weeknd, Kendrick Lamar, Travis Scott, Kesha, Nelly, Tyler, The Creator, Billie Eilish, Metro Boomin, Gwen Stefani, \\ Sean Paul, Vampire Weekend, Jay-Z, Kelly Clarkson, Stevie Wonder, Adele, Arctic Monkeys, Lorde, Britney Spears, Selena Gomez, Daddy Yankee, 21 Savage, \\ David Guetta, J Balvin, The Cure, Bruno Mars, Dua Lipa, Bruce Springsteen, Snoop Dogg, B.o.B, OutKast, Lady Gaga, Hozier, Wiz Khalifa, Foo Fighters, Lana Del Rey, \\ Beyoncé, Madonna, Shakira, John Legend, Mark Ronson, Sam Smith, Billy Joel, Jeremih, Paramore, Chance the Rapper, DJ Snake, Sabrina Carpenter, Kid Cudi, Trey Songz,\\  Kings of Leon, Enrique Iglesias, Pharrell Williams, Arcade Fire, Jessie J, Lil Uzi Vert, Bob Dylan\end{tabular}                                                                                                                                                                                                                                                                                                                                                                                                                                                                                                                                                                                                                                                                                                                                                                                                                                                                                                                                                                                                                                                                                                                                                                              \\ \hline
Actor        & \begin{tabular}[c]{@{}l@{}}Tom Wilkinson, Al Pacino, Kevin Costner, Franco Nero, Philip Seymour Hoffman, Alan Rickman, Leonardo DiCaprio, Ben Affleck, William Hurt, Mark Wahlberg, \\ Jonah Hill, Shia LaBeouf, Don Cheadle, Orlando Bloom, Jeff Goldblum, Denzel Washington, Alec Baldwin, Bradley Cooper, Ed Harris, Jason Clarke, Mahershala Ali, \\ Viggo Mortensen, Owen Wilson, Alan Arkin, James Caan, Nicolas Cage, Samuel L, David Strathairn, Matt Damon, George Clooney, Giovanni Ribisi, Jared Leto, \\ Kevin Spacey, Matthew McConaughey, Gary Sinise, Pete Postlethwaite, Keanu Reeves, Timothy Spall, Harry Dean Stanton, John Carroll Lynch, Chiwetel Ejiofor, \\ Woody Harrelson, Ryan Gosling, Joaquin Phoenix, Donald Sutherland, Paul Dano, Chris Hemsworth, David Oyelowo, Tom Hardy, Barry Pepper, Kurt Russell, \\ Christian Bale, Jeff Daniels, Ben Whishaw, Sterling Hayden, Edward Norton, Sam Shepard, Andy Garcia, Harvey Keitel, Benicio Del Toro, Gene Hackman, Bruce Willis,\\  Guy Pearce, Jonathan Pryce, Michael Fassbender, James Stewart, Zach Galifianakis, Forest Whitaker, Vincent Cassel, Michael Sheen, Tom Berenger, Jim Carrey,\\  Steve Buscemi, Joe Pesci, Christian Berkel, Rutger Hauer, Mel Gibson, Elliott Gould, Tim Robbins, Daniel Craig, Jeffrey Wright, Matthew Modine, Domhnall Gleeson, \\ Brendan Gleeson, John Hurt, Michael Stuhlbarg, Hugo Weaving, John Goodman, Mark Hamill, Colin Farrell, Ken Watanabe, Clint Eastwood, Ralph Fiennes, Val Kilmer, \\ John Hawkes, Ben Kingsley, Seth Rogen, Robert Duvall, Brad Pitt, Max von Sydow, Stanley Tucci, Tom Cruise, Christopher Lloyd, Tommy Lee Jones, Jason Statham, \\ Michael Caine, Paul Giamatti, Josh Hutcherson, Adrien Brody, Michael J, Jeremy Renner, Liam Neeson, Mark Ruffalo, Terrence Howard, John Cleese, Harrison Ford, \\ Clive Owen, Jake Gyllenhaal, Will Smith, Danny DeVito, Elijah Wood, Sean Connery, Tom Sizemore, Stellan Skarsgård, Robin Williams, Hugh Jackman, John Lithgow, \\ Benedict Cumberbatch, Mykelti Williamson, John Malkovich, Gary Oldman, Johnny Depp, Jeff Bridges, Hugh Grant, Jean Reno, Aaron Eckhart, Michael Madsen, \\ Jude Law, J.K, Jon Voight, Casey Affleck, Robert Pattinson, Daniel Brühl, Billy Bob Thornton, Russell Crowe, Ewan McGregor, Christopher Walken, Morgan Freeman, \\ Josh Brolin, Richard Harris, Shea Whigham, Bill Murray, Christoph Waltz, Jamie Foxx, Christopher Plummer, Ethan Hawke, Albert Finney, Miles Teller, Don Johnson, \\ Javier Bardem, Bill Paxton, Robert De Niro, Timothée Chalamet, Sam Rockwell, Kevin Bacon, Simon Pegg, Sean Penn, Ving Rhames, Tom Hanks, Anthony Hopkins, \\ Heath Ledger, Tim Roth, Martin Sheen, Michael Keaton, Joseph Gordon-Levitt, Kyle Chandler, John Travolta, Bruce Dern, Steve Carell, Dustin Hoffman, Oscar Isaac\end{tabular} \\ \hline
Athelete     & \begin{tabular}[c]{@{}l@{}}Dirk Nowitzki, Mia Hamm, Diana Taurasi, Allyson Felix, Sheryl Swoopes, David Ortiz, James Harden, Mike Trout, Mookie Betts, Chris Paul, Aitana Bonmati, \\ Roger Federer, Faker, Annika Sorenstam, Thierry Henry, Jimmie Johnson, Tom Brady, Randy Moss, Shohei Ohtani, Kevin Durant, Zinedine Zidane, Calvin Johnson, \\ Novak Djokovic, LeBron James, Alexia Putellas, Albert Pujols, Max Scherzer, Mariano Rivera, Ichiro Suzuki, Patrick Mahomes, Aaron Donald, Steve Nash, \\ Georges St-Pierre, Giannis Antetokounmpo, Stephen Curry, Floyd Mayweather, Clayton Kershaw, Manny Pacquiao, Andrés Iniesta, Barry Bonds, Tim Duncan, \\ Lauren Jackson, Luka Modric, Shelly-Ann Fraser Pryce, Bryce Harper, Ray Lewis, Simone Biles, Rafael Nadal, Derek Jeter, Shaun White, Michael Schumacher, \\ Peyton Manning, Candace Parker, Nikola Jokic, Serena Williams, Jason Kidd, Andy Murray, Mikaela Shiffrin, Lewis Hamilton, Lisa Leslie, Bernard Hopkins, Kobe Bryant, \\ Justin Verlander, Tamika Catchings, Alex Rodriguez, Jon Jones, Tiger Woods, Dwyane Wade, Kohei Uchimura, Michael Phelps, Xavi Hernandez, Cristiano Ronaldo,\\  Usain Bolt, Max Verstappen, Kawhi Leonard, Venus Williams, Katie Ledecky, Kylian Mbappé, Maya Moore, Alex Ovechkin, Sidney Crosby, Phil Mickelson, \\ Adrian Beltré, Kevin Garnett, Miguel Cabrera, Lionel Messi\end{tabular}                                                                                                                                                                                                                                                                                                                                                                                                                                                                                                                                                                                                                                                                                                                                                                                                                                                                                                                                                                                                                                                                                                                                                                                                                                                                                                                                                                                                                                                                  \\ \hline
Entrepreneur & Brian Chesky, Garrett Camp, Kevin Systrom, Sam Walton, Larry Ellison, Ratan Tata, Ritesh Agarwal, Ted Turner, Steve Jobs, Jack Ma, Richard Branson, Jeff Bezos,                                                                                                                                                                                                                                                                                                                                                                                                                                                                                                                                                                                                                                                                                                                                                                                                                                                                                                                                                                                                                                                                                                                                                                                                                                                                                                                                                                                                                                                                                                                                                                                                                                                                                                                                                                                                                                                                                                                                                                                                                                                                                                                                                                                                                                                                                                                                                                                                                                                                                                                                                                                                                                                                                                         \\ \hline
Politician   & \begin{tabular}[c]{@{}l@{}}Che Guevara, Mike Pence, Li Ka-shing, Woodrow Wilson, Dwight D. Eisenhower, Abdel Fattah el-Sisi, Franklin D. Roosevelt, Yasser Arafat, Haruhiko Kuroda, \\ Donald Trump, Bill Clinton, Stephen Schwarzman, Narendra Modi, Gianni Infantino, Masayoshi Son, Bernard Arnault, Hui Ka Yan, Benjamin Netanyahu, Winston Churchill,\\  Vladimir Putin, John F. Kennedy, Theodore Roosevelt, Nelson Mandela, Sergey Brin, Margaret Thatcher, Xi Jinping, Ronald Reagan, Golda Meir, Recep Tayyip Erdogan, \\ Charles de Gaulle, Jim Yong Kim, Warren Buffett, Qamar Javed Bajwa, Jerome H. Powell, Wang Jianlin, Lech Wałęsa, Michel Temer, Doug McMillon, \\ Mohammed bin Salman Al Saud, Lloyd Blankfein, Lee Hsien Loong, Jawaharlal Nehru, Shinzo Abe, Michael Bloomberg, Tony Blair, Li Keqiang, Rodrigo Duterte, \\ Justin Trudeau, Hu Jintao, Bob Iger, Mario Draghi, Khalifa bin Zayed Al-Nahyan, Bashar al-Assad, Ayatollah Khomeini, Ali Hoseini-Khamenei, Indira Gandhi, \\ Deng Xiaoping, Kim Jong-un, Ma Huateng, Joseph Stalin, Mikhail Gorbachev, Moon Jae-in, Mary Barra, Mahatma Gandhi, Christine Lagarde, Jokowi Widodo, Mao Zedong,\\  Ken Griffin, Mustafa Kemal Atatürk, Theresa May, Aliko Dangote, Darren Woods, Jiang Zemin, Rupert Murdoch, Fidel Castro, Jean-Claude Juncker, Robert Mueller, \\ Enrique Pena Nieto, Carlos Slim Helu, Tim Cook, Robin Li, Antonio Guterres, Larry Fink\end{tabular}                                                                                                                                                                                                                                                                                                                                                                                                                                                                                                                                                                                                                                                                                                                                                                                                                                                                                                                                                                                                                                                                                                                                                                                                                                                                                                                                                                                                                   \\ \hline
Scientist    & \begin{tabular}[c]{@{}l@{}}lbert Einstein, Fei-Fei Li, Jennifer Doudna, Yann LeCun, Thomas Edison, Gregor Mendel, Geoffrey Hinton, John von Neumann, Max Planck, Sam Altman, Tim Berners-Lee, \\ Andrew Ng, Rosalind Franklin, Andre Geim, Marie Curie, Ilya Sutskever, Kip Thorne, James Watson, Srinivasa Ramanujan, Nikola Tesla, Niels Bohr, Enrico Fermi, \\ Rachel Carson, Yoshua Bengio, Alan Turing, Stephen Hawking, Francis Crick, Linus Pauling, Demis Hassabis, Werner Heisenberg, Barbara McClintock\end{tabular}                                                                                                                                                                                                                                                                                                                                                                                                                                                                                                                                                                                                                                                                                                                                                                                                                                                                                                                                                                                                                                                                                                                                                                                                                                                                                                                                                                                                                                                                                                                                                                                                                                                                                                                                                                                                                                                                                                                                                                                                                                                                                                                                                                                                                                                                                                                                         
\end{tabular}}
  \caption{Names for each sub-domain of the celebrity recognition subset.}
  \label{tab:celebrity}
  \vspace{-2mm}
  \end{table*}

\lstdefinelanguage{json}{
    basicstyle=\ttfamily\footnotesize, 
    numbers=left,
    numberstyle=\scriptsize,
    stepnumber=1,
    showstringspaces=false,
    breaklines=true,
    frame=lines,
    backgroundcolor=\color{gray!10},
    morestring=[b]",
    morecomment=[l]{//},
    commentstyle=\color{gray},
    stringstyle=\color{teal}
}

\begin{figure*}[t] 
    \centering    
    \begin{lstlisting}[language=json]
{
    "Instruction": "I will provide you with an image of a person. Please list as many detailed attributes as possible based on the following categories. Below are some examples you can refer to:",
    "Gender": ["Male", "Female", "Unknown"],
    "Age": ["Infant", "Child", "Teenager", "Adult", "Elderly", "Unknown"],
    "Ethnicity": ["Caucasian", "African", "Asian", "...", "Unknown"],
    "Occupation": ["Doctor", "Engineer", "Teacher", "...", "Unknown"],
    "Appearance": {
        "Hair": {
            "Type": ["Short", "Long", "Curly", "Straight", "Bald", "Ponytail", "Unknown"],
            "Color": ["Red", "Blue", "Green", "Yellow", "Black", "White", "...", "Unknown"],
            "Description": ["Short black hair with dotted pattern", "...", "Unknown"]
        },
        "Beard": {
            "Type": ["Full Beard", "Goatee", "Mustache", "Unknown"],
            "Color": ["Red", "Blue", "Green", "Yellow", "Black", "White", "...", "Unknown"],
            "Description": ["Black full beard", "...", "Unknown"]
        },
        "Expression": {
            "Type": ["Smiling", "Frowning", "Surprised", "Angry", "Sleeping", "Crying", "Laughing", "...", "Unknown"],
            "Description": ["Smiling with closed eyes", "...", "Unknown"]
        }
    },
    "Clothing & Accessories": {
        "Clothing": {
            "Upper": {
                "Type": ["T-shirt", "Shirt", "Blouse", "Unknown"],
                "Color": ["Red", "Blue", "Green", "Yellow", "Black", "White", "...", "Unknown"],
                "Description": ["Purple hooded puffer jacket", "...", "Unknown"]
            },
            "Lower": {
                "Type": ["Jeans", "Skirt", "Shorts", "Unknown"],
                "Color": ["Red", "Blue", "Green", "Yellow", "Black", "White", "...", "Unknown"],
                "Description": ["Black pants with dotted patterns", "...", "Unknown"]
            },
            "Shoes": {
                "Type": ["Sneakers", "Boots", "Sandals", "Barefoot", "Unknown"],
                "Color": ["Red", "Blue", "Green", "Yellow", "Black", "White", "...", "Unknown"],
                "Description": ["White sneakers with red stripes", "...", "Unknown"]
            }
        },
        "Accessories": {
            "Hat": {
                "Type": ["Baseball Cap", "Beanie", "Fedora", "Unknown"],
                "Color": ["Red", "Blue", "Green", "Yellow", "Black", "White", "...", "Unknown"],
                "Description": ["Black beanie with white stripes", "...", "Unknown"]
            },
            "Glasses": {
                "Type": ["Sunglasses", "Prescription Glasses", "Goggles", "Unknown"],
                "Color": ["Red", "Blue", "Green", "Yellow", "Black", "White", "...", "Unknown"],
                "Description": ["Black sunglasses with red frame", "...", "Unknown"]
            }
        }
    },
    "Posture": {
        "Type": ["Standing", "Sitting", "Lying", "Arms Crossed", "Arms Raised Overhead", "...", "Unknown"],
        "Description": ["Sitting on a bench", "Standing on one leg", "...", "Unknown"]
    },
    "Actions": {
        "Type": ["Walking", "Running", "Jumping", "Sitting", "Standing", "Sleeping", "...", "Unknown"],
        "Description": ["Walking a dog", "Reading a book with a red cover", "...", "Unknown"]
    }
}
    \end{lstlisting}
    \caption{Prompt used to generate structured property dictionary.}
    \label{fig:structured_prompt}
\end{figure*}

\end{document}